\documentclass[sigconf]{acmart}
\usepackage{multirow}
\AtBeginDocument{%
  }

\copyrightyear{2022}
\acmYear{2022}
\setcopyright{acmcopyright}\acmConference[MM '22]{Proceedings of the 30th ACM
International Conference on Multimedia}{October 10--14, 2022}{Lisboa, Portugal}
\acmBooktitle{Proceedings of the 30th ACM International Conference on Multimedia
(MM '22), October 10--14, 2022, Lisboa, Portugal}
\acmPrice{15.00}
\acmDOI{10.1145/3503161.3548341}
\acmISBN{978-1-4503-9203-7/22/10}

\begin{document}

\title{MVPTR: Multi-Level Semantic Alignment for Vision-Language Pre-Training via Multi-Stage Learning}

\author{Zejun Li}
\orcid{0000-0002-7443-8027}
\email{zejunli20@fudan.edu.cn}
\author{Zhihao Fan}
\email{fanzh18@fudan.edu.cn}
\affiliation{%
  \institution{Fudan University}
  \streetaddress{220 Handan Rd}
  \city{Yangpu Qu}
  \state{Shanghai Shi}
  \country{China}
}

\author{Huaixiao Tou}
\affiliation{%
  \institution{ByteDance}
  \city{Shanghai}
  \country{China}}
\email{zhangyuan.zhang@bytedance.com}

\author{Jingjing Chen}
\orcid{0000-0002-7443-8027}
\email{chenjingjing@fudan.edu.cn}
\author{Zhongyu Wei}
\authornote{Corresponding author.}
\email{zywei@fudan.edu.cn}
\author{Xuanjing Huang}
\email{xjhuang@fudan.edu.cn}
\affiliation{%
  \institution{Fudan University}
  \streetaddress{220 Handan Rd}
  \city{Yangpu Qu}
  \state{Shanghai Shi}
  \country{China}
}

\renewcommand{\shortauthors}{Zejun Li et al.}

\begin{abstract}

Previous vision-language pre-training models mainly construct multi-modal inputs with tokens and objects (pixels) followed by performing cross-modality interaction between them. We argue that the input of only tokens and object features limits high-level semantic alignment like phrase-to-region grounding. Meanwhile, multi-level alignments are inherently consistent and able to facilitate the representation learning synergistically. Therefore, in this paper, we propose to learn \textbf{M}ulti-level semantic alignment for \textbf{V}ision-language \textbf{P}re-\textbf{TR}aining (\textbf{MVPTR}). In MVPTR, we follow the nested structure of both modalities to introduce concepts as high-level semantics. To ease the learning from multi-modal multi-level inputs, our framework is split into two stages, the first stage focuses on intra-modality multi-level representation learning, the second enforces interactions across modalities via both coarse-grained and fine-grained semantic alignment tasks. In addition to the commonly used image-text matching and masked language model tasks, we introduce a masked concept recovering task in the first stage to enhance the concept representation learning, and two more tasks in the second stage to explicitly encourage multi-level alignments across modalities.
\end{abstract}



\maketitle

\begin{figure}
\centering
\includegraphics[width=0.42\textwidth]{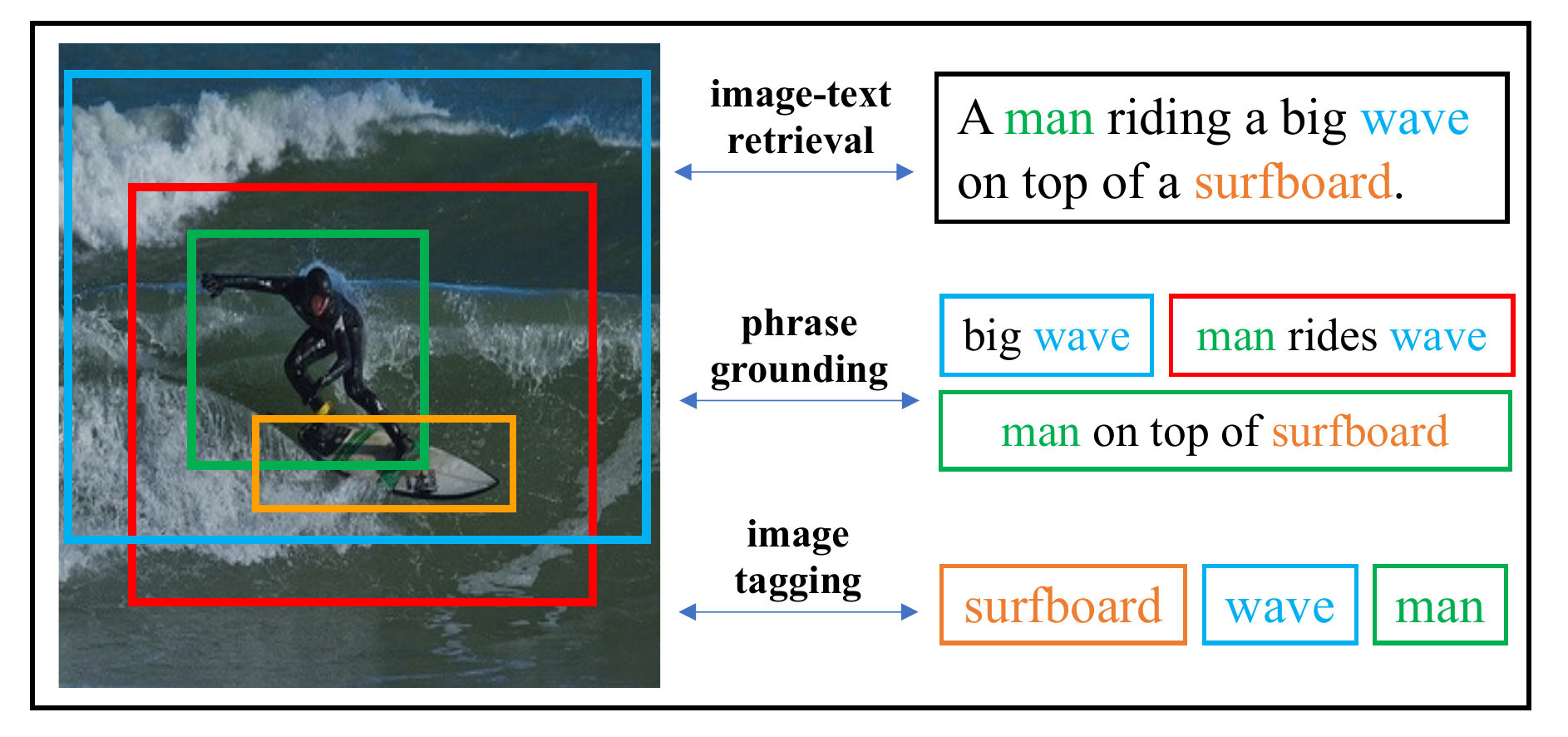}
\centering
\caption{An illustration of multi-level alignments between image and text. Colors indicate the correspondence.}
\label{intro}
\end{figure}

\section{Introduction}
\label{sec:intro}

Language and vision are two important aspects for human to perceive the world. Therefore, it is essential to enable the machine to process multi-modal information for the development of advanced artificial intelligence. Recent years have witnessed extensive research efforts to align the semantics across vision and language with various tasks, e.g., image-text retrieval~\cite{lin2014microsoft,young2014image}, visual question answering~\cite{antol2015vqa,goyal2017making}, and phrase grounding~\cite{yu2016modeling,plummer2015flickr30k}. 
In order to break boundaries between tasks, pre-training methods are developed to learn the cross-modality representations based on large-scale corpora consisting of image-text pairs~\cite{lu2019vilbert,li2019visualbert, su2019vl,li2020oscar,tan2019lxmert,li2020unicoder,zhou2020unified,li2021align,kim2021vilt,zhang2021vinvl} and can be adapted to downstream tasks through fine-tuning. 

Existing vision-language pre-training (VLP) methods use the paradigm of sequence modeling following BERT~\cite{devlin2018bert}. They construct multi-modal sequences from image-text pairs, textual segments are tokenized into tokens and visual tokens are initialized as objects extracted by off-the-shelf detectors~\cite{lu2019vilbert,chen2019uniter,li2020oscar,zhang2021vinvl}, or patches encoded by vision Transformers~\cite{xu2021e2e,kim2021vilt,li2021align}. Cross-modality interaction and alignment are learned between tokens and objects/patches. Although they have achieved state-of-the-art results on vision and language tasks, seldom of them have explored the multi-level semantic alignments beyond the token-to-region level.


 Figure~\ref{intro} presents the landscape of cross-modality semantic alignment at multiple levels. Two key points stand out. Firstly, semantics in both modalities reveal the nested property. On the language side, a word is the basic unit, structurally related words form a phrase, and several phrases make up a sentence. On the vision side, an image is composed of several sub-images (i.e., bounding boxes), bounding boxes naturally contain multi-level semantics as they consist of one or several objects. Secondly, to comprehensively understand image-text pairs, different tasks require semantic alignments at different levels which are mutually reinforcing. Image tagging labels images (objects) with tokens, which helps phrase grounding to link regions and phrases, e.g., grounding the blue box in Figure~\ref{intro} to ``wave'' helps match it to ``a big wave'' and ``man rides wave''. Image-text retrieval aims at global alignment, which benefits from phrase-level and object-level alignment. Conversely, learning high-level alignment helps to refine the low-level ones.

However, the high-level alignment is limited by the input of only tokens and region features. A region might correspond to a phrase with several tokens, they should be connected and considered as a unit but such phrase-level semantics are not explicit in the token sequence. Region features lie in a different space from token embeddings while object-level semantics contained in object tags can help align regions with tokens and phrases. Motivated by these findings, we argue that it is effective to model multi-level semantics for both modalities and enforce cross-modality interactions at different semantic levels for pre-training.

In this paper, we propose to learn \textbf{M}ulti-level semantic alignment for \textbf{V}ision-language \textbf{P}re-\textbf{TR}aining (\textbf{MVPTR}). To enable multi-level alignments, we construct input of different semantic granularity for each modality. On the text side, attribute and relation tuples in scene graphs are introduced as phrase concepts, they serve as intermediate nodes to link related tokens, the encoded phrases aggregate information from related tokens and are better units to interact with regions. On the vision side, we leverage object tags as anchors to align region features with text embeddings.


To facilitate the model to learn from multi-level inputs without being interfered, we disentangle the attention for intra-modal and cross-modal interactions by dividing the framework into two stages, 
namely, uni-modal learning and cross-modal learning. The former aims to learn modality-specific interaction, we introduce the task of masked concept recovering (MCR) to align objects and tag concepts on the vision side, encode structural knowledge of phrases on the language side, forming the basis of multi-modal learning. The second stage enforces the interactions across modalities at different semantic levels. Visual semantic contrastive (VSC) loss is firstly utilized to capture the coarse-grained cross-modality interaction. Then, separately encoded visual and textual sequences are concatenated and fed to a multi-modal encoder. Weakly-supervised phrase grounding (WPG) is adopted for fine-grained phrase-level semantic alignment. Based on previously learned alignments, our model is further optimized on high-level tasks of image-text matching (ITM) and masked language model (MLM). 

Our contribution can be summarized as three-fold:
\begin{enumerate}
    \item We propose a new VLP method MVPTR, where we introduce phrases and object tags as concepts to explicitly model multi-level semantics in image-text pairs. We further design several tasks to encourage multi-level alignments jointly.
    \item To ease the multi-level modeling, we divide our framework into a uni-modal stage and a cross-modal stage, the model is optimized to learn specific-level alignment in two stages.
    \item Experimental results verify the effectiveness of MVPTR on several downstream vision and language tasks. We generate state-of-the-art results with significant improvement on image-text retrieval.
\end{enumerate}

\section{Related Work}


Most VLP methods learn to model the cross-modal interaction and alignment through multi-layer Transformer encoders~\cite{tan2019lxmert,chen2019uniter,li2020oscar,li2020unimo,cui2021rosita,gan2020large,zhang2021vinvl}. According to how to transform image inputs into sequences, existing works can be divided into 2 categories. The first category employs off-the-shelf object detectors~\cite{anderson2018bottom,zhang2021vinvl} to detect all possible objects in the image, further encodes those object-centric regions with convolutional neural networks~\cite{tan2019lxmert,chen2019uniter,li2020oscar,gan2020large,li2020unimo,zhang2021vinvl}. With the recent progress in developing vision Transformers, another line of research employs vision Transformers to extract visual tokens from images directly~\cite{kim2021vilt,li2021align,xu2021e2e}, in which image patches are encoded and considered as visual tokens. Our work belongs to the first category, we aim to explore the multi-level semantics contained in object-centric regions.

According to the interaction schema, previous works can be classified into single-stream and 2-stream approaches. Single-stream methods concatenate textual and visual tokens at input-level, and encode the multi-modal sequence collectively, e.g., UNITER~\cite{chen2019uniter}, and OSCAR~\cite{li2020oscar}. 2-stream models encode intra-modality relationship by uni-modal encoders first (e.g., ViLBERT~\cite{lu2019vilbert}, LXMERT~\cite{tan2019lxmert}). MVPTR is related to 2-stream models, but we propose to learn specific-level semantic alignment to guide the uni-modal learning explicitly. Therefore, we regard our method as a multi-stage model.

Most methods employ multi-modal encoders to measure fine-grained image-text alignment but suffer from high computation costs. To accelerate the inference on image-text retrieval, some researchers explore coarse image-text matching. By learning an aligned global visual semantic embedding (VSE) space from huge-scale in-house corpora, CLIP~\cite{radford2021learning} and ALIGN~\cite{jia2021scaling} achieve remarkable performance with faster inference speed. With public data, LightnongDOT is proposed in~\cite{sun2021lightningdot} as a pre-ranker for fine-grained methods. ALBEF~\cite{li2021align} further unifies coarse and fine-grained alignment in a model. We consider coarse and fine-grained image-text matching as different-level alignments in MVPTR.

The idea of utilizing multi-level semantics in VLP methods is firstly introduced in OSCAR~\cite{li2020oscar} and further extended by VinVL~\cite{zhang2021vinvl}. Both OSCAR and VinVL utilize object-level semantics in tags. In MVPTR, region-to-tag alignment is explicitly learned in the uni-modal stage, we further learns phrase-level semantics to connect related tokens and enable region-to-phrase interaction. 


In MVPTR, structurally related tokens in scene graphs are considered phrases. ERNIE-ViL~\cite{yu2020ernie} and ROSITA~\cite{cui2021rosita} also utilize the structural knowledge in scene graphs, both methods design masking strategies to incorporate knowledge implicitly while we propose to explicitly model that knowledge as learnable concepts.





\section{Method}

Figure~\ref{framework} shows the overall architecture of the proposed framework of MVPTR. It includes two stages for uni-modal and cross-modal learning respectively. We will firstly introduce how to construct multi-level inputs before going into the details about the two stages. 

\subsection{Construction of Multi-Level Input}
\label{method_construct}

To enable multi-level semantic alignment, we introduce intra-modal knowledge and build two-level semantics for both language and vision as input. Phrase concepts and tag concepts are extracted as the high-level semantics of words and objects respectively. We denote the input image-text pairs as $\{(V_i, S_i)\}_{i=1}^n$, where $V_i$ and $S_i$ represent the i-th image and text, respectively. Figure~\ref{input_data} illustrates the process.

\subsubsection{Textual Input} For the textual input, it is first tokenized similar to the process in~\cite{devlin2018bert}, we can get the token sequence $w=\{w_{\text{cls}}, w_1,w_2,...,w_N\}$ where $w_{\text{cls}}$ is the special token ``[CLS]''. All tokens are represented by the summation of their corresponding word embeddings, position embeddings, and segment embeddings.

\begin{figure}
\centering
\includegraphics[width=0.49\textwidth]{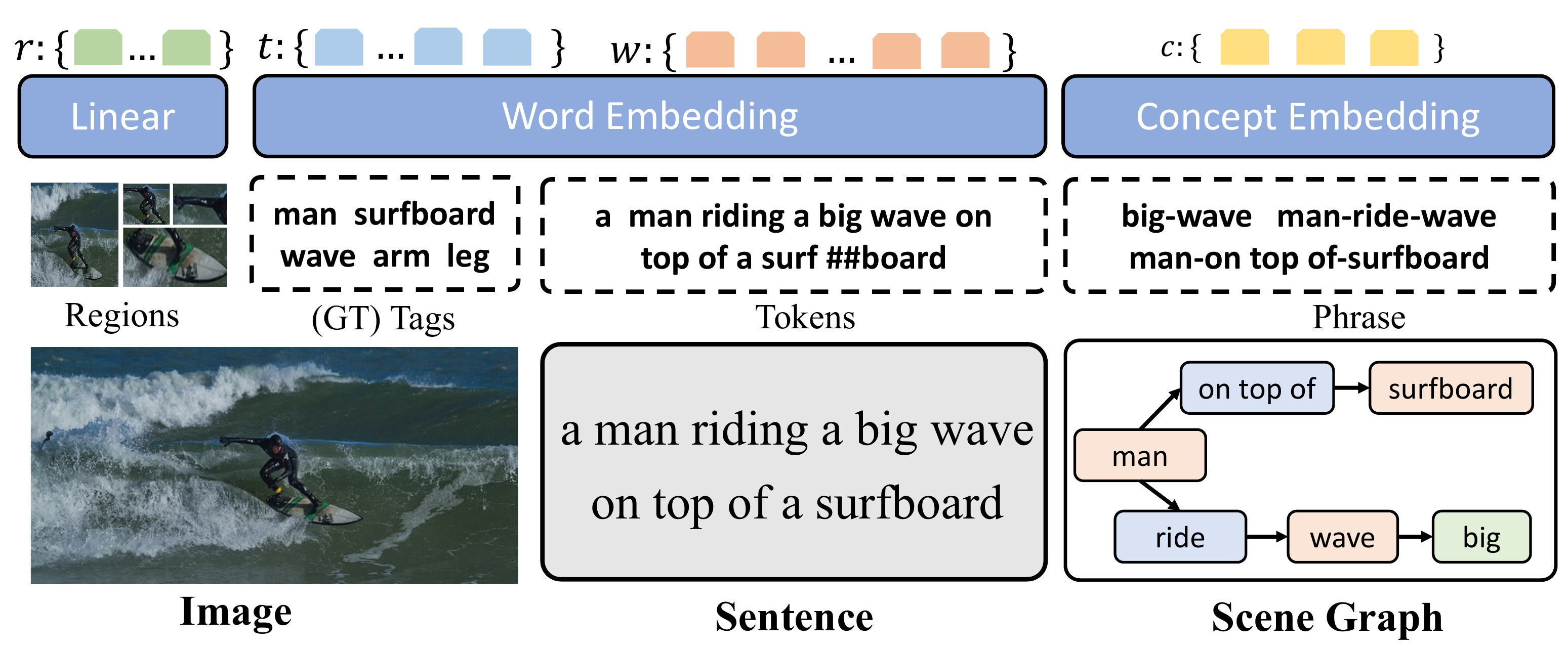}
\centering
\caption{An illustration of the construction of multi-level input from an image-text pair.}
\label{input_data}
\end{figure}

We further construct phrase-level inputs in language. Previous works~\cite{sun2019ernie,yu2020ernie, cui2021rosita} show that structural knowledge contained in texts benefits the representation learning in pre-training. Instead of designing knowledge-guided masking strategies in their works, we consider structurally related tokens as phrases and explicitly learn the knowledge as concepts. Phrase concepts connect related tokens like nouns and their modifiers. Phrase-level representations learned from them contain rich information to further interact with regions.  Specifically, we employ the off-the-shelf parser~\cite{anderson2016spice} to construct scene graphs from sentences. Scene graphs consist of structural tuples of objects, attributes, and relations, where we consider attribute and relation tuples as phrases (e.g., ``big wave'', ``man ride wave'', and ``man on top of surfboard'' in Figure~\ref{input_data}). We learn a concept embedding for each phrase that appears in at least 50 sentences in the training corpus to ensure generalizability. To enable phrase concepts to connect related tokens, we initialize the embeddings of phrase concepts as the average embedding of its tokens. All extracted and valid phrases are denoted as $c=\{c_1,c_2,...,c_M\}$. 

The input textual sequence ${w,c}$ consists of tokens and phrase concepts. It will be fed to the textual encoder where phrase concepts serve as bridges to connect related tokens, and the learned uni-modal phrase representations can be further aligned with regions.


\begin{figure*}
\centering
\includegraphics[width=0.90\textwidth]{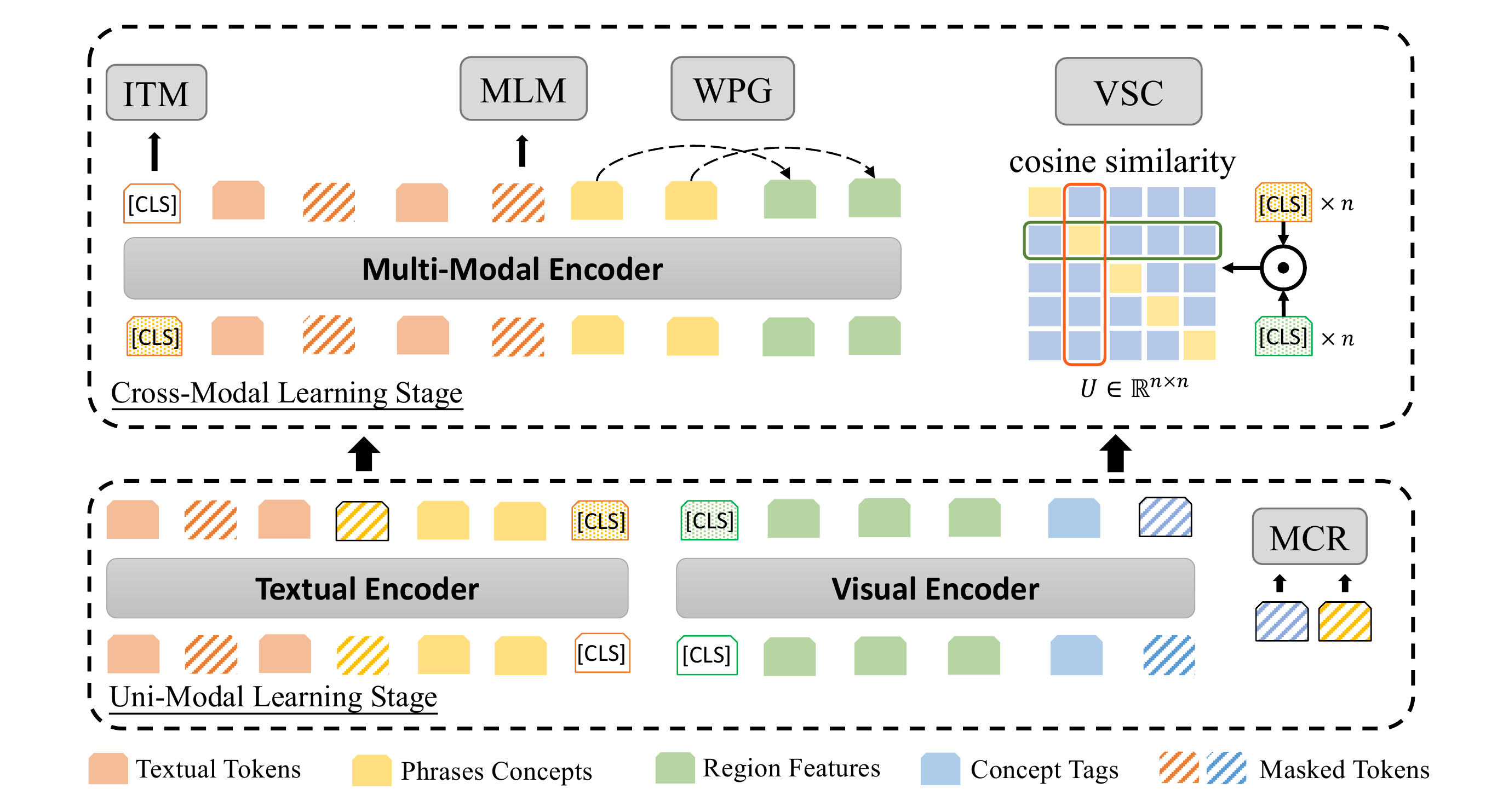}
\centering
\caption{The framework of MVPTR. Acronyms in grey boxes represent the pre-training tasks: masked concept recovering (MCR), visual semantic contrastive (VSC) learning, weakly-supervised phrase grounding (WPG), masked language model (MLM), and image-text matching (ITM). VSC loss is illustrated for a batch of $n$ image-text pairs, $U$ denotes the in-batch similarity matrix, the red and green boxes respectively represent the image-to-text and text-to-image similarities in Equation~\ref{t2i_distribution}. Dashed lines indicate the weakly-supervised setting for WPG.}
\label{framework}
\end{figure*}

\subsubsection{Visual Input} For the input image, an object detector is used to extract object regions from the image, represented by a sequence of position-aware features $r=\{r_1, r_2, ..., r_K\}$, where $r_i \in \mathbb{R}^{2054}$ is a concatenation of the CNN encoded region feature and normalized bounding box positions and sizes. Feature vectors in $r$ are further linearly mapped to ensure the same dimension for all input vectors.

As region features $r$ lie in a different semantic space from token and phrase embeddings, some objects are not mentioned in the paired text, it is difficult to directly align regions with tokens and phrases. We introduce object tags  $t=\{t_1,t_2,...,t_K\}$ as object-level concepts in the visual input. Object tags are inherently consistent with region features as shown in~\cite{li2020oscar}, easing the object-level alignment. At the same time, tags share the same embeddings with tokens, in order to bridge the gap before the cross-modal interaction is learned. The coverage of tags benefits the representation learning of unmentioned objects. Object tags can be straightforwardly obtained from the class names predicted by the object detector. But for datasets with ground truth object tags like MSCOCO~\cite{lin2014microsoft} and OpenImages~\cite{kuznetsova2020open}, we directly utilize those labels as better object-level concepts to guide the visual representation learning.

The input visual sequence fed to the visual encoder is $\{w_{\text{cls}}, t, r\}$. Grounding regions to the corresponding object tags incorporates object-level semantics into the encoded region features, which benefits the cross-modal alignment between regions and tokens/phrases.


\subsection{Uni-Modal Learning}

As we construct multi-level inputs from images and texts. It is hard for merged attention to learn both cross-modal and intra-modal interaction appropriately. We divide the encoder into 2 stages.

In the uni-modal stage, we restrict the cross-modal interaction, textual and visual inputs are separately encoded. Textual sequence $\{w,c\}$ is fed to the textual encoder while $\{t,r\}$ is encoded by the visual encoder. Corresponding outputs are denoted as $\Tilde{w}, \Tilde{c}, \Tilde{r}, \Tilde{t}$. Different from previous 2-stream methods~\cite{lu2019vilbert,tan2019lxmert}, we design an MCR (masked concept recovering) task to guide the uni-modal representation learning with the information contained in concepts. The uni-modal stage is able to learn modality-specific interactions. On the language side, structural information contained in phrases are encoded and contextual phrase representations are learned. On the vision side, MCR grounds region features to tag concepts, which serve as anchors to bridge 2 modalities and benefit the further phrase-level alignment. Generally, the uni-modal stage forms the basis of the multi-modal interaction and alignment.



\subsubsection{Masked Concept Recovering}
\label{mcr_method}

We first illustrate the MCR task on the visual side. Following BERT~\cite{devlin2018bert}, input object tags $t$ are randomly masked (replaced by ``[MASK]'' or a random token or stay unchanged). Since the structure of tag sequences is simpler than that of captions, we increase the masking probability from 0.15 to 0.25 for higher training efficiency. We denote the partially masked sequence as $\hat{t}$ and the masked token set as $\hat{t}^\text{m}$. With the output of visual encoders, for each masked concept, an MLP with softmax is employed to predict a probability distribution on all candidate concepts: $p^{\text{mcr}}(\hat{t}_i^\text{m}\vert r,\hat{t}\setminus\{\hat{t}^\text{m}\})$. Different from MLM in BERT, concepts are disordered, when 2 tags ``cat dog'' are both masked, we should not restrict the prediction to the original order, i.e. the same loss for ``cat dog'' and ``dog cat''. Therefore we resort to using Hungarian matching~\cite{stewart2016end, carion2020end, hu2021vivo} to reorder the labels:
\begin{align}
    \hat{\alpha} = \arg\min_{\alpha}\sum_{i=1}^{\Vert \hat{t}^\text{m}\Vert}1-p^{\text{mcr}}(\hat{t}_{\alpha(i)}^{\text{m}}\vert r,\hat{t}\setminus\{\hat{t}^\text{m}\}) \nonumber \\
    \text{s.t.} \;\; \alpha(i)\neq\alpha(j)\;\; \text{if}\; i\neq j
    \label{hung}
\end{align}
Solving~\ref{hung} yields the linear optimal labels-to-predictions assignment $\hat{\alpha}$. MCR objective on the visual side is formulated as:
\begin{align}
    \mathcal{L}_{\text{mcr}}^v=\frac{1}{\Vert \hat{t}^\text{m}\Vert} \sum_{i=1}^{\Vert \hat{t}^\text{m}\Vert}\text{H}(y^\text{mcr}_{\hat{\alpha}(i)},p^{\text{mcr}}(\hat{t}_{\hat{\alpha}(i)}^\text{m}\vert r,\hat{t}\setminus\{\hat{t}^\text{m}\})) 
\end{align}
where $\text{H}$ denotes cross entropy and $y^\text{mcr}_{\hat{\alpha}(i)}$ represents a one-hot distribution of the $\hat{\alpha}(i)$-th original concept.

Similarly, MCR is also applied on the textual encoder as:
\begin{align}
    \mathcal{L}_{\text{mcr}}^s=\frac{1}{\Vert \hat{c}^\text{m}\Vert} \sum_{i=1}^{\Vert \hat{c}^\text{m}\Vert}\text{H}(y^\text{mcr}_{\hat{\beta}(i)},p^{\text{mcr}}(\hat{c}_{\hat{\beta}(i)}^\text{m}\vert w,\hat{c}\setminus\{\hat{c}^\text{m}\})) 
\end{align}
where $\hat{c}^{\text{m}}$ are the randomly masked phrases, $p^{\text{mcr}}(\hat{c}_{\hat{\beta}(i)}^\text{m}\vert w,\hat{c}\setminus\{\hat{c}^\text{m}\})$ is the predicted probability on all candidate phrases on top of the textual encoder outputs, $\hat{\beta}$ is the reordered phrase index mapping similar to $\hat{\alpha}$ in Equation~\ref{hung}, $y^\text{mcr}_{\hat{\beta}(i)}$ is the one-hot label.

As pointed out in~\cite{li2021align}, masked modeling aims to maximize the mutual information between the masked and unmasked sequences.

MCR on the visual encoder can be regarded as weakly-supervised learning for the alignment between region features and tag concepts, analogous to the image tagging task. As tags come from the textual embedding space, they serve as anchors to bridge the 2 modalities. Grounding region features to their tags facilitate the alignment between regions and tokens/phrases in the cross-modal stage.

MCR on the textual side requires the model to reconstruct the structure of the scene graph. It naturally injects structural knowledge into the textual encoder.





\subsection{Cross-Modal Learning}

Based on the uni-modal outputs, the second stage of MVPTR models the cross-modal interaction. In this stage, we model interactions from shallow to deep, encouraging the alignments from phrase-to-region level to text-to-image level.  We first consider coarse image-text matching using visual semantic contrastive (VSC) loss to further align uni-modal outputs. Then, weakly-supervised phrase grounding (WPG) is adopted for phrase-level semantic alignment. Based on previously learned interaction, image-text matching (ITM) grasps the global understanding to perform fine-grained image-text alignment. Masked language model (MLM) is used to further optimize the model to perform high-level reasoning.

\subsubsection{Visual Semantic Contrastive Loss}

This objective aims to align the global image and text embeddings into a shared visual semantic space. We take the L2-normalized output representations of ``[CLS]'' tokens from uni-modal textual and visual encoders as global embeddings of the images and texts. Similar to CLIP~\cite{radford2021learning} and ALBEF~\cite{li2021align}, we consider all $n\times n$ pairs in a training batch of $n$ image-text pairs $\{(V_i, S_i)\}_{i=1}^n$, the in-batch cosine similarity matrix can be computed as $U=\Tilde{T}_{\text{cls}}^\top \Tilde{W}_{\text{cls}}$, where $\Tilde{T}_{\text{cls}}=[\Tilde{t}_{\text{cls}}^1, \Tilde{t}_{\text{cls}}^2,...,\Tilde{t}_{\text{cls}}^n]\in \mathbb{R}^{d \times n}, \Tilde{W}_{\text{cls}}=[\Tilde{w}_{\text{cls}}^1, \Tilde{w}_{\text{cls}}^2,...,\Tilde{w}_{\text{cls}}^n]\in \mathbb{R}^{d\times n}$ are the global embeddings of $n$ images and $n$ sentences, respectively. 

In-batch image-to-text and text-to-image similarities are derived by normalizing $S$ with softmax:
\begin{align}
    p^{\text{v2s}}(V_i,S_k)=\frac{\exp(U_{i,k}/\tau)}{\sum_{j=1}^n \exp(U_{i,j}/\tau)}\;\; \forall \;1\leq k \leq n \nonumber \\ p^{\text{s2v}}(S_j,V_k)=\frac{\exp(U_{k,j}/\tau)}{\sum_{i=1}^n \exp(U_{i,j}/\tau)} \;\; \forall \;1\leq k \leq n \label{t2i_distribution}
\end{align}
where $\tau$ is a learnable temperature parameter, considering only the matched pairs as positive samples (the yellow diagonal in the similarity matrix of Figure~\ref{framework}), the contrastive learning applies a cross entropy loss on the predicted similarity as:
\begin{align}
    \mathcal{L_{\text{vsc}}}=-\frac{1}{2n}(\sum_{i=1}^n \log p^{\text{v2s}}(V_i,S_i)+\sum_{j=1}^n \log p^{\text{s2v}}(S_j,V_j)) 
\end{align}



VSC encourages the shallow interaction to coarsely align images and texts, it further narrows the gap between uni-modal outputs before the deep interaction is learned.

With the help of the first stage and VSC loss, an aligned multi-modal sequence $\{\Tilde{w},\Tilde{c},\Tilde{r}\}$ is constructed. Note that tags on the vision side are excluded to avoid shortcuts in masked language model.

\subsubsection{Weakly-supervised Phrase Grounding}

Having grounded regions to their object-level concepts and learned phrase-level concepts in the uni-modal stage, we are able to explicitly encourage phrase-level semantic alignment. Without ground-truth correspondence, we propose to use weakly-supervised phrase grounding (WPG). We take L2-normalized representations of the phrases and regions encoded by the multi-modal encoder, denoted as $\{g(\Tilde{c}_i)\}_{i=1}^M$ and $\{g(\Tilde{r}_j)\}_{j=1}^K$, then we aggregate phrase-to-region similarities into text-to-image similarity:
\begin{align}
    S^{\text{wpg}}(V,S)=\frac{1}{M}\sum_{i=1}^M \max_{j=1}^K g(\Tilde{c}_i)^\top g(\Tilde{r}_j)
\end{align}
this equation is based on the assumption of multiple instance learning~\cite{karpathy2015deep}, behind which the object-region alignment learned in MCR helps,  then hinge loss is used:
\begin{align}
    \mathcal{L}_{\text{wpg}}=\frac{1}{n}\sum_{i=1}^n \min(0, 0.2+S^{\text{wpg}}(V_{i'}, S_i)-S^{\text{wpg}}(V_{i}, S_i))
\end{align}
where $(V_i,S_i)$ is the positive pair, and $V_{i'}$ is a hard negative image sampled from $p^{\text{s2v}}(S_j,V_k)$ in Equation~\ref{t2i_distribution}.

We only consider text-to-image similarity in the training objective, since object-centric methods tend to detect all possible objects, some of which may be noisy or not mentioned in the corresponding text, this phenomenon may lead to unreliable image-to-text similarity measurement. Motivated by~\cite{li2020does}, WPG is applied on the 3-rd layer of the 6-layer multi-modal encoder.

As phrase concepts contain richer information than tokens and tags, WPG benefits the visual representation learning. Furthermore, phrase-level alignment provides information for ITM and MLM.

\subsubsection{Image Text Matching}

Based on previous object-level, coarse-level, and phrase-level alignments, ITM trains the model to perform fine-grained image-text matching. With the help of the uni-modal learning stage, it is possible to sample in-batch hard negative pairs from the distributions $p^{\text{v2s}(V_i,S_k)}, p^{\text{s2v}}(S_j,V_k)$ defined in Equation~\ref{t2i_distribution}. ITM is formulated as a binary classification task, taking the multi-modal encoder's output representation of the ``[CLS]'' token as the global representation of the image-text pair, an MLP with softmax is learned to predict a binary distribution on if the pair matches: $p^{\text{itm}}(V,S)$, ITM loss is defined as:
\begin{align}
    \mathcal{L}_{\text{itm}} = \frac{1}{\Vert B\Vert} \sum_{(V,S)\in B}\text{H}(y^{\text{itm}}(V,S),p^{\text{itm}}(V,S))
\end{align}
where $B$ is a batch of $n$ matched and $n$ hard negative pairs. $y^{\text{itm}}(V,S)$ indicates whether $(V,S)$ is a positive pair.

\subsubsection{Masked Language Modeling}

Besides multi-level alignments, MLM with visual clues is the key in VLP methods, we apply it in the second stage. Similar to MCR, the textual token sequence is randomly masked, we use $\hat{w}$ and $\hat{w}^{\text{m}}$ to denote the masked sequence and masked token set. For each masked token, an MLP with softmax is learned to predict a probability distribution over all candidate tokens: $p^{\text{mlm}}(\hat{w}_i^\text{m}\vert r,c,\hat{w}\setminus\{\hat{w}^\text{m}\})$. MLM aims to minimize:
\begin{align}
    \mathcal{L}_{\text{mlm}}=\frac{1}{\Vert \hat{w}^\text{m}\Vert} \sum_{i=1}^{\Vert \hat{w}^\text{m}\Vert}\text{H}(y^\text{mlm}_i,p^{\text{mlm}}(\hat{w}_i^\text{m}\vert r,c,\hat{w}\setminus\{\hat{w}^\text{m}\}))
\end{align}
where $y_i^{\text{mlm}}$ is the one-hot distribution of the original token. 

In MVPTR, MLM is regarded as a high-level inference task when tokens like adjectives, nouns, verbs are masked. Both ITM and MLM are applied on top of the whole model to utilize and refine the previously learned alignments and interaction.



\subsection{Multi-Stage Pre-training}

To ease the cross-modal learning. We further divide the pre-training procedure into 2 stages. In the first stage, intra-modal interaction (MCR) and coarse image-text alignment (VSC) are learned. Uni-modal encoders are trained to generate pre-aligned visual and textual representations by optimizing the objective:
\begin{align}
    \mathcal{L}_1=\mathcal{L}_{\text{mcr}}^v+\mathcal{L}_{\text{mcr}}^s+\mathcal{L}_{\text{vsc}}
\end{align}
In the second stage, the cross-modal encoder is considered. Our complete model is trained to minimize the full objective:
\begin{align}
    \mathcal{L}_2=\mathcal{L}_{\text{mcr}}^v+\mathcal{L}_{\text{mcr}}^s+\mathcal{L}_{\text{vsc}}+\mathcal{L}_{\text{mlm}}+\mathcal{L}_{\text{itm}}+\mathcal{L}_{\text{wpg}}
\end{align}

\begin{table*}[]
\caption{Fine-tuned results of base-size vision-language pre-trained methods on downstream tasks. Bold numbers represent the best performance in each column.}
    \centering
    \resizebox{0.90\textwidth}{!}{
\begin{tabular}{l|ccccccc|cccc}
\toprule
\multicolumn{1}{c|}{\multirow{3}{*}{\begin{large}Method\end{large}}} & \multicolumn{7}{c|}{MSCOCO (5K test set)}                                                           & \multicolumn{2}{c}{VQA}                               & \multicolumn{2}{c}{SNLI-VE}                  \\ \cmidrule{2-8} \cmidrule{8-12} 
\multicolumn{1}{c|}{}                        & \multicolumn{3}{c}{Caption Retrieval} & \multicolumn{3}{c}{Image Retrieval} & \multirow{2}{*}{RSUM} & \multirow{2}{*}{test-dev} & \multirow{2}{*}{test-std} & \multirow{2}{*}{val} & \multirow{2}{*}{test} \\
\multicolumn{1}{c|}{}                        & R@1         & R@5        & R@10       & R@1        & R@5        & R@10      &                       &                           &                           &                      &                       \\ \midrule
UNITER~\cite{chen2019uniter}                                       & 65.7        & 88.6       & 93.8       & 52.9       & 79.9       & 88.0      & 468.9                 & 72.70                     & 72.91                     & 78.59                & 78.28                 \\
OSCAR~\cite{li2020oscar}                                        & 70.0        & 91.1       & 95.5       & 54.0       & 80.8       & 88.5      & 479.9                 & 73.16                     & 73.44                     & -                    & -                     \\
VILLA~\cite{gan2020large}                                        &             & -          &            &            & -          &           & -                     & 73.59                     & 73.67                     & 79.47                & 79.03                 \\
UNIMO~\cite{li2020unimo}                                        &             & -          &            &            & -          &           & -                     & 73.79                     & 74.02                     & 80.00                 & 79.10                  \\
ROSITA~\cite{cui2021rosita}                                       & 71.3        & 91.6       & 95.6       & 54.4       & 80.9       & 88.6      & 482.4                 & 73.91                     & 73.97                     & -                    & -                     \\
ALBEF (4M)~\cite{li2021align}                                        & 73.1        & 91.4       & 96.0       & 56.8       & 81.5       & 89.2      & 488.0                 & 74.54                     & 74.70                     & 80.14                & \textbf{80.30}                 \\
VinVL~\cite{zhang2021vinvl}                                        & 74.6        & 92.6       & 96.3       & 58.1       & 83.2       & 90.1      & 494.9                 & 75.95                     & 76.12                     & -                    & -                     \\
MVPTR (Ours)                                         & \textbf{77.3}        & \textbf{93.6}       & \textbf{96.9}       & \textbf{60.1}       & \textbf{84.0}       & \textbf{90.7}      & \textbf{502.6}                 & \textbf{76.16}                     &    \textbf{76.36}                       & \textbf{80.30}                 & 80.17                  \\ \bottomrule
\end{tabular}}
    \label{full_res}
\end{table*}

\begin{table}[]
\caption{Image-text retrieval results on Flickr30k. Uni-modal encoder based methods encode images and texts separately and use cosine similarities. We fine-tune CLIP (ViT-B-32) and denote it as ``CLIP*'', ``MVPTR-Uni'' denotes the uni-modal stage of our method.}
\resizebox{0.49\textwidth}{!}{
\begin{tabular}{lcccccc}
\toprule
\multicolumn{1}{c}{\multirow{3}{*}{Method}} & \multicolumn{6}{c}{Flickr (1K test set)}                                                       \\
\multicolumn{1}{c}{}                        & \multicolumn{3}{c}{Caption Retrieval}          & \multicolumn{3}{c}{Image Retrieval}           \\
\multicolumn{1}{c}{}                        & R@1           & R@5           & R@10          & R@1            & R@5           & R@10          \\ \midrule
UNITER                                      & 85.9          & 97.1          & 98.8          & 72.5           & 92.4          & 96.1          \\
VILLA                                       & 86.6          & 97.9          & 99.2          & 74.7           & 92.9          & 95.8          \\
UNIMO                                       & 89.7          & 98.4          & 99.1          & 74.7          & 93.4          & 96.1          \\
ALBEF (4M)                                       & 94.3          & 99.4          & 99.8          & 82.8           & 96.7          & 98.4          \\
MVPTR                                        & \textbf{95.2} & \textbf{99.7} & \textbf{100}  & \textbf{84.0}  & \textbf{96.8} & \textbf{98.5} \\ \midrule
\multicolumn{7}{l}{\textbf{\textit{Uni-Modal Encoder based Methods}}}                                                                                               \\
LightningDot                                & 83.9          & 97.2          & 98.6          & 69.9           & 91.1          & 95.2          \\
CLIP*                                      & 87.2          & 98.1          & 99.6          & 71.6           & 91.9          & 96.0          \\
MVPTR-Uni                                        & \textbf{87.4} & \textbf{98.2} & \textbf{99.7} & \textbf{74.5} & \textbf{94.0} & \textbf{96.9} \\ \bottomrule
\end{tabular}}
\label{fk_itr}
\end{table}

\section{Experiments}

\subsection{Implementation Details}
\label{pretrain_set}


We employ the architecture of a 6-layer Transformer encoder for all encoders, following the base setting in previous works~\cite{lu2019vilbert,chen2019uniter,li2020oscar,gan2020large,li2020unimo,cui2021rosita,zhang2021vinvl,li2021align}. The textual encoder and visual encoder are both initialized from the first 6 layers of BERT-base, while the multi-modal encoder is initialized from the last 6 layers.

We pre-train our model on a large-scale vision-language corpus, including MSCOCO~\cite{lin2014microsoft}, Flirck30k~\cite{young2014image}, GQA~\cite{hudson2019gqa}, Conceptual Captions~\cite{sharma2018conceptual}, SBU~\cite{ordonez2011im2text} and OpenImages~\cite{kuznetsova2020open}. We also exclude the val/test splits of Flickr30K provided by~\cite{karpathy2015deep}, which is used in VinVL pre-training. For images, we adopt the object detector provided in ~\cite{zhang2021vinvl} to extract object features and tags. The maximal lengths for textual tokens, phrase concepts, regions feature sequence, and object tags are set to 35, 5, 50, and 20, respectively.

Our model is trained with AdamW~\cite{loshchilov2017decoupled} optimizer and batch size of 1024. The first training stage consists of 15 epochs while the second one has 30 epochs, the peak learning rates are both set as 1e-4 and warmed-up from 0 in the first 1/5 of the process, then linearly decay to 0. The supplementary file provides more details.

\subsection{Downstream Tasks}


\subsubsection{Image-Text Retrieval}

Image-to-caption and caption-to-image retrieval require the model to judge the semantic similarities between image-text pairs. We fine-tune our model on Flickr30K~\cite{young2014image} and MSCOCO~\cite{lin2014microsoft} with the split provided by~\cite{karpathy2015deep}. We rely on the coarse and fine-grained image-text alignment ability of MVPTR brought by VSC and ITM. The model is fine-tuned with $\mathcal{L}_{\text{vsc}}+\mathcal{L}_{\text{itm}}$. During inference, we employ a re-ranking strategy: for each image (sentence), the uni-modal encoders are used to compute coarse similarities, and only the $k_c (k_i)$ most similar captions (images) are re-ranked by the multi-modal encoder. In practice, we set $k_c=128,k_i=64$. We also compute the RSUM metric as the sum of
recall metrics at K = \{1, 5, 10\} of both caption and image retrieval, RSUM reflects the overall performance.

\subsubsection{Multi-Modal Classification}

We adapt our pre-trained models to 2 widely-used classification tasks on image-text pairs, visual question answering (VQA), and visual entailment (VE). For VQA, we evaluate our model on VQA v2~\cite{goyal2017making}, Following~\cite{zhang2021vinvl}, the task is formulated as a classification task to pick an answer from a shared answer set (3129 candidates). 

For the VE task on SNLI-VE~\cite{xie2019visual}, we follow~\cite{chen2019uniter} to consider it as a 3-way classification, which infers whether the relationship between an image-text pair is entailment, neutral, or contradictory.

On both tasks, we take the jointly encoded representation of the ``[CLS]'' token and employ a task-specific MLP with softmax to predict the probabilities of candidate choices.


\begin{table}[]
    \centering
    \caption{Phrase Grounding Results on Flickr30k Entities. Results of VisualBERT are provided in~\cite{dou2021improving}.}
    \begin{tabular}{lcc}
\toprule
\multirow{2}{*}{Method} & \multicolumn{2}{c}{Flirck30k Entities} \\ \cmidrule{2-3} 
                        & Pre-trained        & Fine-tuned        \\ \midrule
VisualBERT~\cite{li2019visualbert}              & 35.53              & 71.33             \\
VinVL                   & 24.08              & -                 \\
MVPTR                   & \textbf{42.55}     & \textbf{73.10}    \\ \bottomrule
\end{tabular}
    \label{fk_ent_table}
\end{table}

\begin{table*}
\caption{Ablation studies. We use the notation in Section~\ref{method_construct} where $w,c,r$, and $t$ respectively denote tokens, phrases, region features, and tags. ``Uni-RSUM'' denotes the RSUM of uni-modal encoder based methods. ``MergAttn'' is short for ``merged attention'' where the uni-modal stage is removed and a 12-layer Transformer is adopted to model the full sequence $\{w,c,r,t\}$.}
\begin{tabular}{llcccccc}
\toprule
\multirow{2}{*}{Inputs} & \multirow{2}{*}{Training Tasks} & \multicolumn{2}{c}{MSCOCO 5k} & VQA &  \multicolumn{2}{c}{Flickr30k Entities} \\
                              &                                 & RSUM         & Uni-RSUM       & test-dev           & Pre-trained         & Supervised        \\ \midrule
$w,c,r,t$                       & $\text{MCR}^{v,s}$+VSC+WPG+MLM+ITM             & 502.6        & 471.9          & 76.16                 & 42.55               & 73.10             \\
$w,c,r,t$                       & $\text{MCR}^{s}$+VSC+WPG+MLM+ITM             & 498.1             &    463.2            &  75.57              & 32.05              &  72.86                               \\
$ w,c,r $                         & $\text{MCR}^s$+VSC+WPG+MLM+ITM             & 494.8              & 459.5               & 74.44                 &  27.60            & 71.98                               \\
$ w,c,r,t $                       & $\text{MCR}^{v}$+VSC+MLM+ITM                 & 498.3              & 468.8               & 76.02               & 36.09             & 72.65                              \\
$ w,r,t $                         & $\text{MCR}^v$+VSC+MLM+ITM                 & 500.4             &    470.9            & 75.98                  & 35.04             & 72.48                               \\
$w,c,r,t$ &  $\text{MCR}^{v,s}$+WPG+MLM+ITM (MerAttn)                                 & 496.5                  &    NA               & 75.25                            &  29.42              & 72.24 \\
\bottomrule
\end{tabular}
\label{ablation}
\end{table*}

\begin{figure*}
\centering
\includegraphics[width=0.94\textwidth]{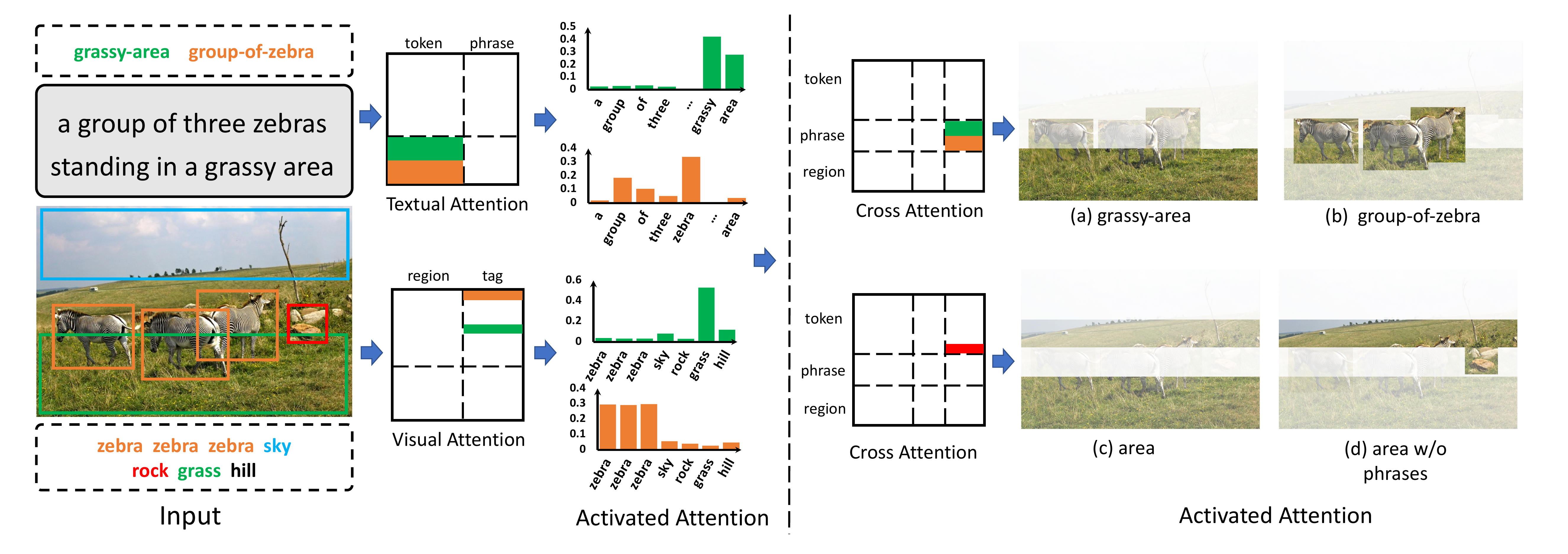}
\centering
\caption{Visualization of the learned attention in our textual, visual, and multi-modal encoders inspired by~\cite{cui2021rosita}. The left part shows the phrase-to-token and region-to-tag attention in uni-modal encoders. The right part illustrates the phrase/token-to-region attention in the multi-modal encoder, (a,b,c) come from MVPTR while (d) comes from a variant of row 5 in Table~\ref{ablation}.}
\label{vis_attention}
\end{figure*}
\subsubsection{Visual Grounding}

Visual phrase grounding requires models to perform phrase-level alignment. We evaluate our model on Flickr30k Entities~\cite{plummer2015flickr30k}, which considers contextual phrases in a sentence. We follow~\cite{dou2021improving} to take the cosine similarities between the mean pooling of phrase tokens and region candidates as the phrase-to-region similarities. IoU (Intersection of Union) scores between detected bounding boxes and ground-truth boxes are adopted as soft labels to fine-tune the similarities. We ground a phrase to the box with the highest similarity. Following~\cite{rohrbach2016grounding}, a prediction is correct if its IoU is greater than 0.5. Results of the pre-trained and fine-tuned models are reported.


\subsection{Overall Performance}

\subsubsection{Evaluation on Image-Text Retrieval}
Table~\ref{full_res} and Table~\ref{fk_itr} shows the retrieval results on two datasets. The full MVPTR achieves a significant improvement on both MSCOCO and Flickr30k. On MSCOCO, MVPTR even outperforms ALIGN~\cite{jia2021scaling} that trains on a corpus of 1.8B images and achieves RSUM of 500.4. This result shows the learned fine-grained image-text alignment with the assist of other-level alignment.

As we propose to learn aligned uni-modal representations to ease cross-modal learning. We evaluate the uni-modal encoders on Flickr30k in Table~\ref{fk_itr}. Compared with the other 2 pre-trained methods using uni-modal encoders, LightningDot~\cite{sun2021lightningdot} and CLIP~\cite{radford2021learning}, our uni-modal encoders perform better with about 50\% parameters (6 layers versus 12 layers), the advance is consistent on MSCOCO which is listed in the supplementary file. The improvement is attributed to the anchor role of tags and MCR that enhances it.
 
\begin{figure*}
\centering
\includegraphics[width=0.9\textwidth]{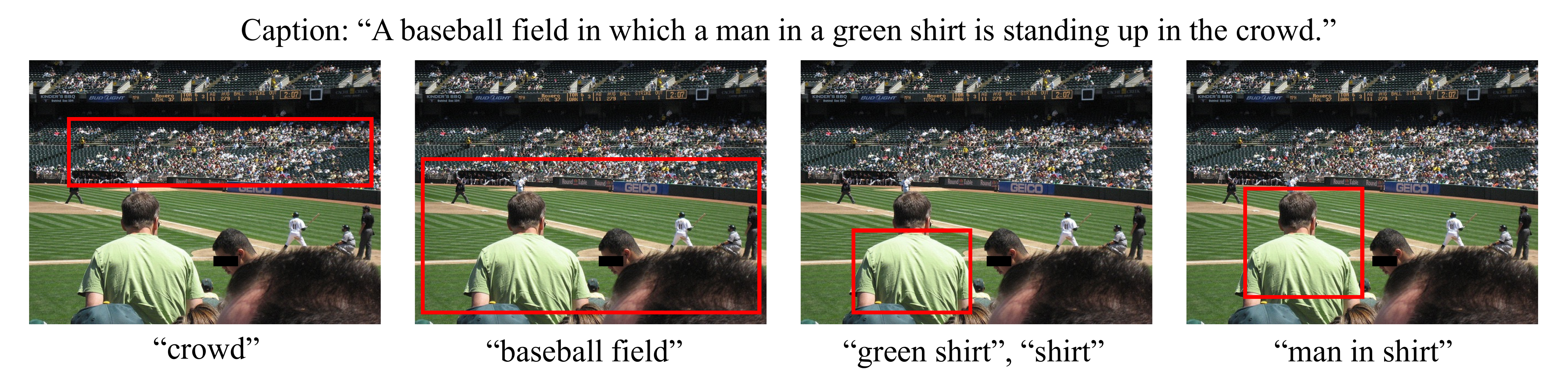}
\centering
\caption{Visualization of the phrase grounding on an unseen image-text pair. For each noun entity and extracted phrase, the red box indicates the detected region with the highest similarity with it.}
\label{case}
\end{figure*}

\subsubsection{Evaluation on VQA and VE}
As for VQA, MVPTR outperforms all other pre-trained models. In the detailed comparison with VinVL, the advantage of our model is more significant in the type of ``other'' questions (0.67 versus 0.66), in which general reasoning ability is required~\cite{goyal2017making}. As the interaction between tags and questions is prohibited in MVPTR pre-training, we believe our model learns the appropriate cross-modal interaction and reasoning. More detailed results of VQA are provided in the supplementary file.

As for VE, MVPTR achieves the state-of-the-art on the validation set but performs slightly worse than ALBEF on the test set. 

\subsubsection{Evaluation on Visual Grounding}
Following~\cite{dou2021improving}, we evaluate the phrase grounding ability of the pre-trained embeddings and fine-tuned ones. With the help of explicit modeling of phrase-level semantics and alignment, MVPTR is stronger at connecting related words to match contextual phrases to regions they describe, especially when no phrase-to-region supervision is provided. 

\subsection{Effectiveness of Multi-Level Alignment}

To validate the effectiveness of introduced multi-level semantic representations and alignment. We first provide quantitative results of ablation studies in Table~\ref{ablation}, the first line represents the full MVPTR model. Notice that if a task is ablated, it will be not used in both training stages. We further provide qualitative results by probing the learned attention for an unseen image-text pair in Figure~\ref{vis_attention}.

\subsubsection{Object-Level Alignment}

By comparing the first row and the third row of Table~\ref{ablation}, we observe that the absence of object-level concepts hinders the model from learning alignment at other levels, especially on coarse image-text matching and phrase grounding. This finding shows that object-level semantics are the foundation of other high-level alignment and interaction. Compared with the implicit manner (row 2), explicit $\text{MCR}^{v}$ (row 1) enhances the alignment, leading to the improvement on all tasks. 

From the bottom left part of Figure~\ref{vis_attention} we can see that the visual encoder successfully aligns region features with tag concepts (e.g, the green box with ``grass''), building the basis for further alignment.

\subsubsection{Phrase-Level Alignment}

As illustrated in the fourth line and the fifth line of Table~\ref{ablation}, removing phrase-level semantics and alignment mainly affects the visual grounding task and fine-grained image-text matching. This phenomenon firstly indicates the effectiveness of providing phrase-level information on visual representation learning, then corresponds to our argument that global alignment can be inferred from the phrase-level alignment. Another finding is that phrase-level semantics may interfere with the representation learning if not explicitly modeled through WPG and $\text{MCR}^s$, which validates the necessity of explicit modeling on the relationship between phrases tokens/regions.

Referring to the top right part of Figure~\ref{vis_attention}, phrase concepts serve as the intermediate nodes to connect related tokens and are able to gather phrase-level information. The visualized attention of ``grassy-area'' and ``group-of-zebra'' over all regions validates the effectiveness of the learned phrase-level interaction. Comparing (c) and (d), we find it difficult for the generic word ``area'' to ground to the correct region without the help of structural information in phrases (both the hill and grassland can be regarded as areas). This problem is alleviated in (c) where intermediate phrase nodes exist.

\subsection{Further Analysis}
\label{fur_anal}

\begin{table}[]
\caption{Statistics of phrases in different datasets. ``Phr.'' is short for phrases.}
\resizebox{0.49\textwidth}{!}{
\begin{tabular}{lccl}
\toprule
Datasets  & \# of Unique Phr. & \# of Phr. per Text & Frequent Phr. \\ \midrule
MSCOCO    & 32311                  & 1.966                    & \begin{tabular}[c]{@{}l@{}}young-man, two-man\\
city-street, white-plate \\ group-of-people\end{tabular}                 \\ \midrule
Flickr30k & 21067                  & 2.188                    & \begin{tabular}[c]{@{}l@{}}two-man, young-boy\\
black-dog, red-shirt \\ man-sit on-bench\end{tabular}                 \\ \midrule
VQA       & 19905                  & 0.828                    & \begin{tabular}[c]{@{}l@{}}many-people, what-color\\
sign-say, man-hold \end{tabular}                  \\ \midrule
SNLI-VE   & 22779                  & 0.8957                    & \begin{tabular}[c]{@{}l@{}}man-ride-bike, little-girl\\
two-man, man-stand \\
man-wear-shirt\end{tabular}                  \\ \bottomrule
\end{tabular}
}
\label{dataset_stat}
\end{table}



\subsubsection{Two-Stage Framework} According to the 7-th row of Table~\ref{ablation}, the variant contains additional multi-level inputs and alignment supervision, but the performance deteriorated when compared with VinVL. Combined with the positive results of MVPTR, we verify our argument in Section~\ref{sec:intro} that it is difficult for merged attention to learn both intra-modal and cross-modal interactions at the same time. The multi-modal stage also benefits from the well-aligned uni-modal representations as illustrated in Table~\ref{fk_itr}. 

\subsubsection{Statistics of Phrases} Table~\ref{dataset_stat} provides the statistics of phrases in downstream datasets. We can see that the learned phrases are quite general across datasets. Each sentence contains about 2 phrases. Common phrases of the main objects like ``two-man'' and ``little girl'' are frequent across different datasets. In VQA, the number of phrases per question is small due to the lengths of questions, but phrases like ``what-color'' provide useful information of the question type. We also find that the words used in hypothesis texts of SNLI-VE are quite different from those of image descriptions, which limit the learned phrases to generalize. As MVPTR only uses the frequency of phrases to ensure generalizability, we will explore to construct more general phrases in the future.

\subsubsection{Visualization of Phrase Representation}

In Figure~\ref{case}, we provide an example to show the learned contextual phrase representations by grounding them to the most similar regions. Our pre-trained model is able to match them to the objects they depict. 


\section{Conclusion}

In this paper, we configure multi-level semantics for language and vision for pre-training. A two-stage pre-training framework is proposed to enforce uni-modal and cross-modal representation learning where multi-level alignment is learned. The proposed model achieves state-of-the-art results on downstream tasks.

\begin{acks}
This work is partially supported by Ministry of Science and Technology of China (No.2020AAA0106701).
\end{acks}

\bibliographystyle{ACM-Reference-Format}
\bibliography{mm22}

\appendix

\section{Additional Implementation Details}

\subsection{Details of Input Construction}

As for the visual input, the confidence threshold set for object extraction is 0.2. Following~\cite{zhang2021vinvl}, each object-centric region is represented by a dense position-aware embedding of 2054 dimensions, the first 2048 dimensions come from the CNN encoder of the object detector, the last 6 dimensions are $[\frac{x_{\text{min}}}{w}, \frac{y_{\text{min}}}{h}, \frac{x_{\text{max}}}{w}, \frac{y_{\text{max}}}{h}, \frac{(x_{\text{max}}-x_{\text{min}})}{w}, \frac{(y_{\text{max}}-y_{\text{min}})}{h}]$ in which  $(x_{\text{min}}, y_{\text{min}})$ and $(x_{\text{max}}, y_{\text{max}})$ are respectively the coordinates of the top left and bottom right points, $w$ and $h$ represent the width and height of the bounding box.

\subsection{Details of Pre-Training}




\subsubsection{Model Architecture}
In our experiments, the pre-trained MVPTR is set to the base setting following~\cite{devlin2018bert}. Specifically, each Transformer layer has 12 attention heads, the hidden size is 768, the intermediate size of the feedfoward layer is set to 3072. The visual, textual, and multi-modal encoders are all stacked 6 Transformer layers. 

\subsubsection{Pre-training Details}
During the pre-training process, the dropout probabilities of hidden feature and attention maps are both set to 0.1. All the pre-training process is performed with 8 NVIDIA TESLA V100 GPUs and the framework of deepspeed~\cite{rasley2020deepspeed}. We employ the ZERO-2 optimization and fp16 to improve the training efficiency, the full process takes around 5 days.

\subsection{Details of Fine-Tuning}

Here we provide more details of the fine-tuning process on downstream tasks. For all tasks, we utilize the AdamW~\cite{loshchilov2017decoupled} optimizer with weight decay and linear learning rate decay scheduler.

\subsubsection{Fine-Tuning on Image-Text Retrieval}

For image-text retrieval, we evaluate the model on MSCOCO~\cite{lin2014microsoft} and Flickr30k~\cite{young2014image}, both datasets accompany each image with five descriptions. Following~\cite{li2021align}, we adopt the split provided by~\cite{karpathy2015deep}, where MSCOCO has around 113k/5k/5k images in the train/validation/test splits and Flickr30k contains 29k/1k/1k images in the train/validation/test splits. The model is fine-tuned for 15 epochs on MSCOCO and 5 epochs on Flickr30k, the batch size is set as 96 per GPU, the peak learning rate is set as 4e-5. The maximal lengths of object regions, tokens, object tags, and phrases are 50, 50, 30, and 5, respectively. The model is trained with a drop-out probability of 0.1 and weight decay of 0.05. We report recall@K (the proportion of queries for which at least one of the top-K retrieved items is labeled positive, $\text{K} \in \{1,5,10\}$) for both image-to-text and text-to-image retrieval.

\subsubsection{Fine-Tuning on VQA}
For the evaluation of VQA, we adopt the VQA v2~\cite{goyal2017making}, which is based on the MSCOCO images. In VQA v2, the training/validation/test sets are respectively made of 83k/41k/81k images. Following previous works~\cite{tan2019lxmert,chen2019uniter, zhang2021vinvl, li2021align}, we fine-tune MVPTR on the training and validation sets to achieve the best performance. We consider the confidence defined in VQA labels as the soft probability labels and utilize binary cross entropy loss. To fully utilize the inference ability brought by masked language model (MLM) in pre-training, we carefully initialize the classification head of VQA. For each candidate answer, the corresponding weight vector and bias in the VQA linear layer are initialized as the average of the weights and biases in the MLM head, respectively. The model is fine-tuned for 25 epochs with a batch size of 32 per GPU, a learning rate of 5e-5, a drop-out probability of 0.3, and weight decay of 0.05, but we find that the model converges in around 7 epochs. The maximal lengths of object regions, tokens, object tags, and phrases are 50, 128, 30, and 5, respectively. We evaluate the accuracy.

\subsubsection{Fine-Tuning on SNLI-VE}
We also evaluate the model on SNIL-VE~\cite{xie2019visual}, which consists of 29783, 1000, and 1000 images in the training, validation, and test sets, respectively. It is formulated as a 3-way classification problem. MVPTR is fine-tuned for 5 epochs with a learning rate of 4e-5, weight decay of 0.05, a drop out probability of 0.1, a batch size of 64 per GPU. The maximal lengths of object regions, tokens, object tags, and phrases are 70, 70, 20, and 5, respectively. The evaluation metric is accuracy.

\subsubsection{Fine-Tuning on Phrase Grounding}
For the visual grounding task on Flickr30k Entities~\cite{plummer2015flickr30k}. We employ the intersection of union (IoU) scores between the predictions and ground truths as soft labels. We have tried 2 settings, the first one considers IoU scores as binary probabilities and employ cross entropy as the objective, the second considers it as a regression problem to use MSE to minimize the differences between the predicted similarities and IoU scores. We empirically find the second setting performs better. We report the accuracy as the ratio of phrases for which the predicted box overlaps with the ground-truth box by more than 0.5 IOU, which is equivalent to Recall@1 used in some papers.

\begin{table}[]
    \centering
    \caption{Image-text retrieval results on MSCOCO 5k test. ``MVPTR-Uni'' represents the uni-modal encoders of the pre-trained MVPTR. We fine-tune the CLIP model with ViT-B/32 and denote it as ``CLIP*''.}
    
\begin{tabular}{l|llllll}
\toprule
\multirow{2}{*}{Model} & \multicolumn{3}{c}{Caption Retrieval}                                        & \multicolumn{3}{c}{Image Retrieval}                                          \\
                       & \multicolumn{1}{c}{R@1} & \multicolumn{1}{c}{R@5} & \multicolumn{1}{c}{R@10} & \multicolumn{1}{c}{R@1} & \multicolumn{1}{c}{R@5} & \multicolumn{1}{c}{R@10} \\ \midrule
LightningDot           & 60.1                    & 85.1                    & 91.8                     & 45.8                    & 74.6                    & 83.8                     \\
CLIP*                   & 63.2                        & 86.6                        & 92.7                         &  44.9                       & 72.1                        & 81.7                         \\
MVPTR-Uni                  & \textbf{68.2}           & \textbf{90.3}           & \textbf{95.0}            & \textbf{51.5}           & \textbf{79.3}           & \textbf{87.6}            \\ \bottomrule
\end{tabular}
    \label{uni-coco}
\end{table}

\begin{table}[]
    \centering
    \caption{Detailed results on the test-std split of VQA v2.}
    \begin{tabular}{lcccc}
\toprule
\multirow{2}{*}{Method} & \multicolumn{4}{c}{VQA v2}                                \\ \cmidrule{2-5} 
                        & yes/no & number         & other          & overall        \\ \hline
VinVL                   & 91.47  & \textbf{61.04} & 66.21          & 76.12          \\
MVPTR                   & \textbf{91.65}  & 58.45 & \textbf{67.16} & \textbf{76.36} \\ \bottomrule
\end{tabular}
    \label{add_vqa}
\end{table}

\section{Additional Results}

\subsection{Additional Results on MSCOCO}
Table~\ref{uni-coco} provides the results of uni-modal encoder based methods on MSCOCO. Compared with two uni-modal encoder based methods, CLIP~\cite{radford2021learning} and LightningDot~\cite{sun2021lightningdot}, the uni-modal encoders of our pre-trained MVPTR show a superior performance on the MSCOCO, which is consistent to the results on Flickr30k.

\subsection{Detailed Results of VQA}
We provide detailed results of different categories in VQA v2. We compare VinVL~\cite{zhang2021vinvl} and our method on the test-std split of VQA v2, the results are provided in Table~\ref{add_vqa}. MVPTR outperforms VinVL on all categories except the ``number'' questions. As the ``number'' questions require the model to count the target objects, we hypothesis the advantage of VinVL mainly comes from the interaction between questions and object tags during pre-training. At the same time, VinVL is pre-trained for around 116 epochs, which is 2.5 times that of MVPTR. More training steps generate more distinct training samples of MLM (with different masked tokens), which provide richer supervision signals to guide the model to learn cross-modal inference, thus improving the performance on tasks like VQA. 

\subsection{Comparison with ALBEF}

ALBEF contains more trainable parameters than MVPTR because that it employs a 12-layer vision Transformer as its visual encoder (a 6-layer Transformer in MVPTR), while other components of the two models are aligned. ALBEF further employs momentum distillation to boost its performance, doubling each training step's computation cost. Under this setting, MVPTR outperforms ALBEF on other tasks while achieving an accuracy of 80.17 on the test set of SNLI-VE, which is comparable to ALBEF's 80.3 and better than its 79.77 without momentum distillation. 

\section{Additional Discussion}
\subsection{Statistics of Phrases}

We provide some statistics of the valid phrases in the pre-training corpus in Table~\ref{phrase_stat}. We have extracted 54950 valid phrases that appear in at least 50 sentences. Compared with the original vocabulary of 30522 words, the introduction of phrases enriches the learned visual concepts. The number of phrases per image/text shows the coverage of phrase concepts. The mean and median frequencies of all phrases illustrate the generalizability of phrases.

Referring to the frequent phrases in the corpus, we find that those phrases are mainly related to the main objects or actions which can be grounded to regions to help the representation learning.

\begin{table}[]
\caption{Statistics of valid phrases in the pre-training corpus.}
\begin{tabular}{lc}
\toprule
\multicolumn{2}{l}{\textit{\textbf{Statistics of Phrases}}}                                                                                               \\
\# All Phrases                & 14.8M                                                                                                                       \\
\# Unique Phrases                & 54950                                                                                                                       \\
Mean/Median Frequency               & 269.38/98 (sentences)                                                                                                                 \\
\# Phrases per Image        & 2.74                                                                                                                        \\
\# Phrases per Text         & 1.58                                                                                                                        \\ \midrule
\multicolumn{2}{l}{\textit{\textbf{Examples of Frequent Phrases}}}                                                                                          \\
\multirow{2}{*}{Attributes} & \multicolumn{1}{l}{group-of-people,  blue-sky,}                                                                         \\
                            & \multicolumn{1}{l}{large-building, little-girl...}                                                                      \\
Relations                   & \multicolumn{1}{l}{\begin{tabular}[c]{@{}l@{}}man-in-shirt,  man-sit at-table\\ artist-perform on-stage\end{tabular}} \\ \bottomrule
\end{tabular}
\label{phrase_stat}
\end{table}


\subsection{Visualization of Non-Contextual Concepts}

\begin{figure}
\centering
\includegraphics[width=0.4\textwidth]{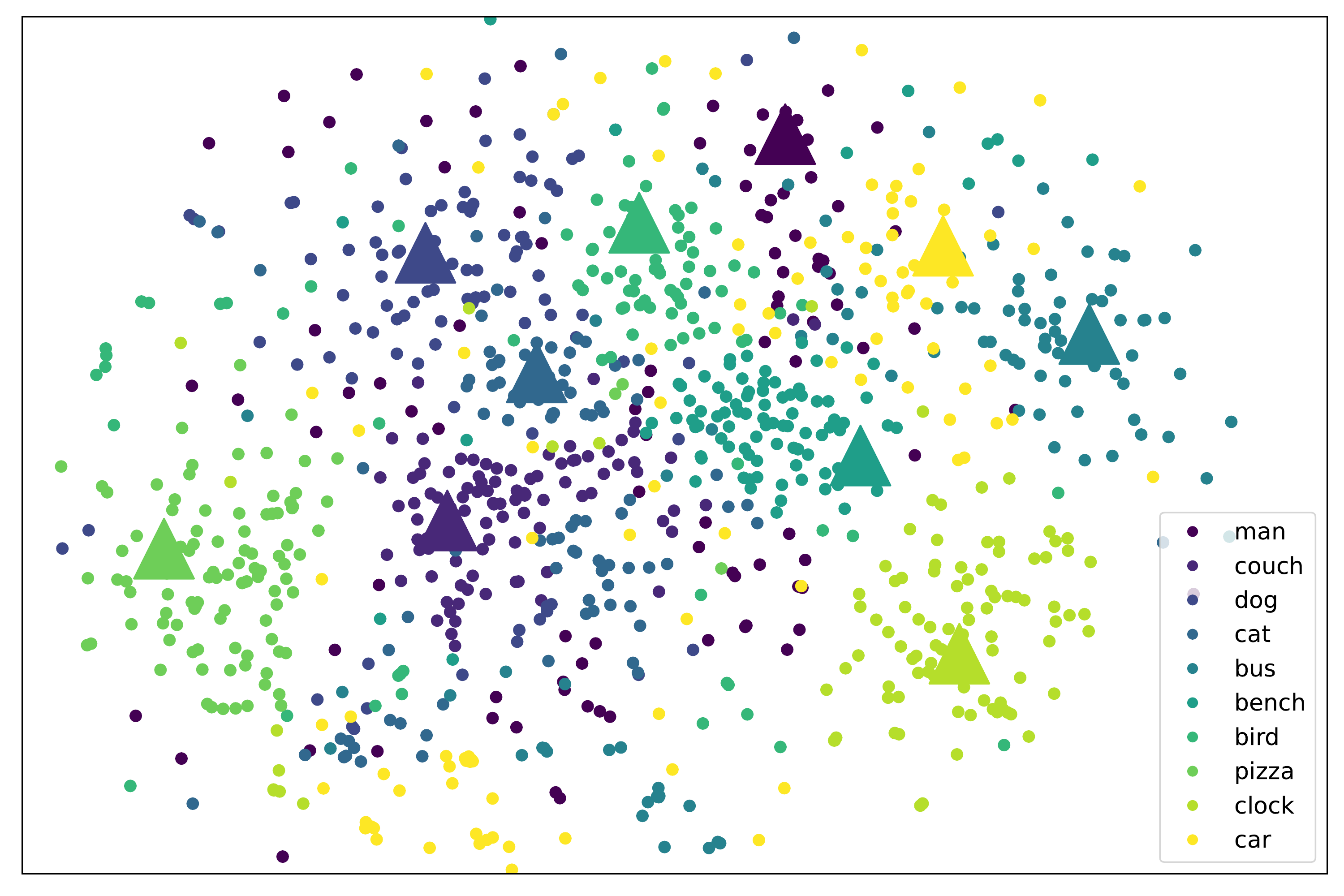}
\centering
\caption{Visualization of learned concepts: triangles represent object concepts and points are related phrase concepts.}
\label{concept_vis}
\end{figure}

We visualize the learned non-contextual embeddings of concepts (input embeddings) with t-SNE in Figure~\ref{concept_vis}. We select common tags (triangles) and phrases (points) that contain those tags. We can observe that phrases are located around the centers of corresponding image tags, showing the hierarchy of the learned semantics. Phrases that include ``man'' or ``car'' have a divergent distribution, which indicates that those 2 kinds of objects always interact with other objects in different scenes. Objects like ``clock'' and 'pizza' appear in specific scenes, the distributions of their related phrases are centralized. ``Couch'', ``dog'', and ``cat'' are closely related, corresponding to the their frequent co-occurrence in home scenes.

\subsection{Extra Computation Cost}

The introduced multi-level semantic alignment will lead to extra computation cost. We empirically find WPG+MCR+VSC brings 3.87\% more pre-training time, which is close to the theoretical 3.4\% more FLOPs (the detailed derivation is omitted here). 

\subsection{Future Works}
Here we list several future directions:
\begin{enumerate}
    \item Better hierarchical modeling of image-text pairs. MVPTR assumes the hierarchy of the image naturally exists among the detected regions, while explicit modeling of object relation can be further detected and modeled.
    \item End-to-end models encode fixed-size patches in vision Transformers or grids in CNN feature maps as the visual tokens, where the basic object-level semantics are absent in the visual sequence. As multi-level information (single objects, regions, and relations) plays a significant role in cross-modal understanding, modeling multi-level semantic representations and alignment in that kind of models is promising.
    \item In MVPTR, phrase concepts represent linguistic knowledge in scene graphs (with the help of SPICE) and object concepts come from the pre-trained visual models (object detectors), those concepts are therefore limited by these tools. We think concepts can be more general and easily extended to a wider area, furthermore, well-learned concepts can also be related to common sense and knowledge graphs.
\end{enumerate}

\end{document}